\documentclass[sigconf]{acmart}

\AtBeginDocument{%
  }

\usepackage[ruled, vlined]{algorithm2e}
\usepackage{multirow}
\usepackage{subfig}

\usepackage{graphicx}
\newcommand{\et}[2]{${#1}^{\pm{#2}}$}

\newcommand{\ourapproach}{SynTalker}

\setcopyright{acmlicensed}
\copyrightyear{2024}
\acmYear{2024}
\acmDOI{10.1145/3664647.3680847}

\acmConference[MM'24]{In Proceedings of the 32nd ACM International Conference on Multimedia}{October 28 - November 1, 2024}{Melbourne, Australia.}
\acmISBN{979-8-4007-0686-8/24/10}

\acmSubmissionID{604}

\begin{document}

\title{Enabling Synergistic Full-Body Control in Prompt-Based Co-Speech Motion Generation}


\author{Bohong Chen}
\orcid{0009-0007-1036-7737}
\affiliation{%
  \institution{State Key Lab of CAD\&CG\\Zhejiang University}
  \city{Hangzhou}
  \state{Zhengjiang}
  \country{China}
}
\email{bohongchen@zju.edu.cn}

\author{Yumeng Li}
\orcid{0009-0007-6558-4165}
\affiliation{%
  \institution{State Key Lab of CAD\&CG\\Zhejiang University}
  \city{Hangzhou}
  \state{Zhengjiang}
  \country{China}
}
\email{yumeng.li@zju.edu.cn}

\author{Yao-Xiang Ding}
\authornote{Corresponding author}
\orcid{0000-0001-8580-1103}
\affiliation{%
  \institution{State Key Lab of CAD\&CG\\Zhejiang University}
  \city{Hangzhou}
  \state{Zhengjiang}
  \country{China}
}
\email{dingyx.gm@gmail.com}

\author{Tianjia Shao}
\orcid{0000-0001-5485-3752}
\affiliation{%
  \institution{State Key Lab of CAD\&CG\\Zhejiang University}
  \city{Hangzhou}
  \state{Zhengjiang}
  \country{China}
}
\email{tjshao@zju.edu.cn}

\author{Kun Zhou}
\orcid{0000-0003-4243-6112}
\affiliation{%
  \institution{State Key Lab of CAD\&CG\\Zhejiang University}
  \city{Hangzhou}
  \state{Zhengjiang}
  \country{China}
}
\email{kunzhou@acm.org}

\renewcommand{\shortauthors}{Bohong Chen, Yumeng Li, Yao-Xiang Ding, Tianjia Shao \& Kun Zhou}

\begin{abstract}
Current co-speech motion generation approaches usually focus on upper body gestures following speech contents only, while lacking supporting the elaborate control of synergistic full-body motion based on text prompts, such as {\it talking while walking}. The major challenges lie in 1) the existing speech-to-motion datasets only involve highly limited full-body motions, making a wide range of common human activities out of training distribution; 2) these datasets also lack annotated user prompts. To address these challenges, we propose {\it \ourapproach}, which utilizes the off-the-shelf text-to-motion dataset as an auxiliary for supplementing the missing full-body motion and prompts. The core technical contributions are two-fold. One is the multi-stage training process which obtains an aligned embedding space of motion, speech, and prompts despite the significant distributional mismatch in motion between speech-to-motion and text-to-motion datasets. Another is the diffusion-based conditional inference process, which utilizes the separate-then-combine strategy to realize fine-grained control of local body parts. Extensive experiments are conducted to verify that our approach supports precise and flexible control of synergistic full-body motion generation based on both speeches and user prompts, which is beyond the ability of existing approaches. Our code, pre-trained models, and videos are available at https://robinwitch.github.io/SynTalker-Page/.
\end{abstract}


\begin{teaserfigure}
  \centering
  \includegraphics[width=0.39\textwidth]{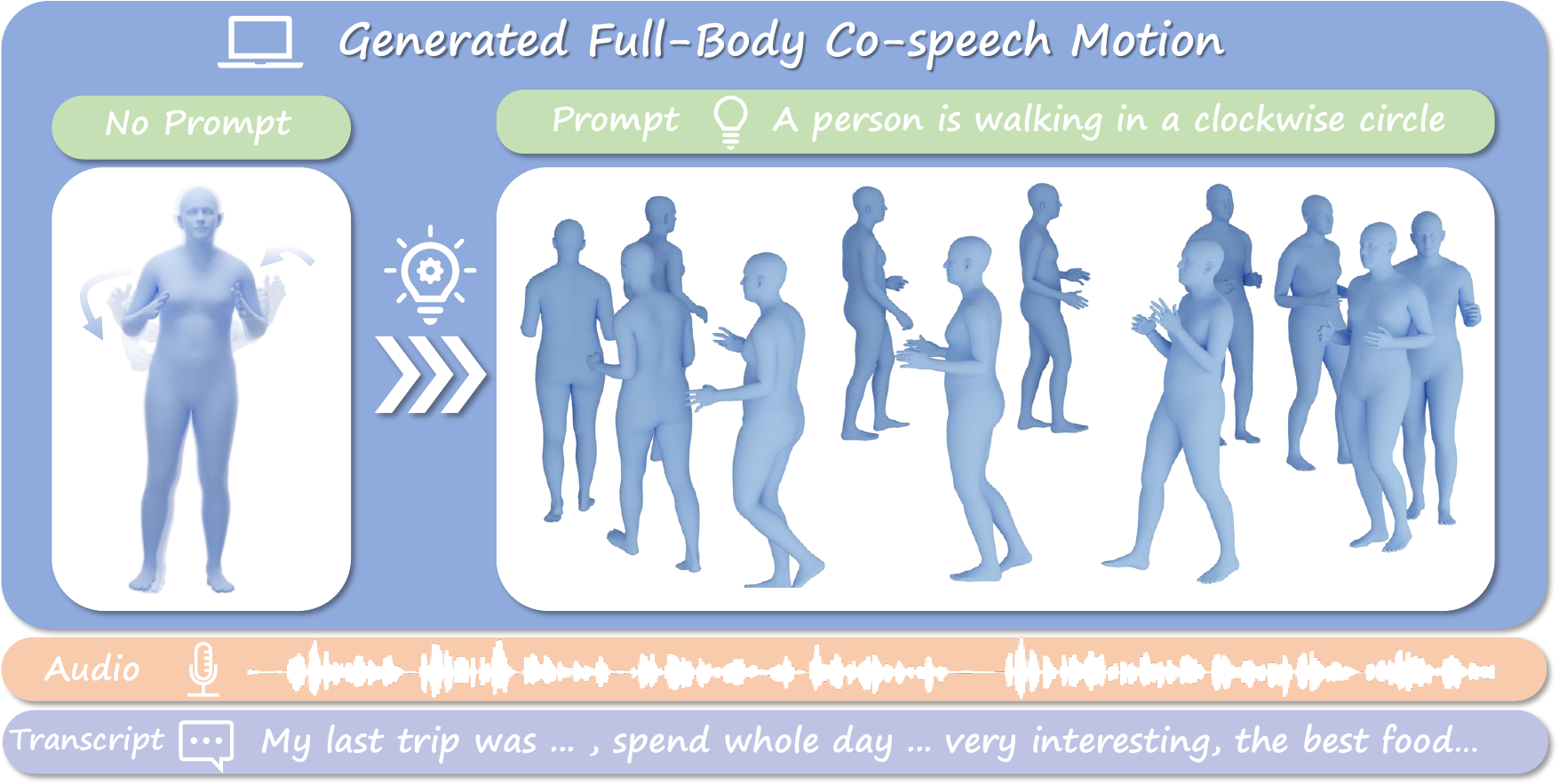}
  \caption{
  Given an audio of speech, as well as \emph{arbitrary motion-related text prompt}, our method can generate full-body synergistic motion matching both speech content and prompt 
  even if the motion is unseen in the speech-to-motion dataset used for training, such as the "walking in a clockwise circle" example in the figure. Meanwhile, the generation result is also highly consistent with the script content and the audio rhythm of the input speech. 
  }
  \label{fig:teaser}
\end{teaserfigure}


\maketitle

\section{Introduction}
Co-speech motion generation~\cite{ghorbani2022zeroeggs,liu2022beat,zhu2023taming,yang2023diffusestylegesture,chhatre2023emotional,EMOTE}, which generates stylized movements of human body following speech audio inputs, is among the central tasks in creating digital talking avatars. Though growing rapidly in recent years, current co-speech motion generation approaches usually focus on upper-body gestures, such as head and hands, or only support limited full-body motions, in special restricted low-body movements. One of the fundamental challenges here is that the speech signal is too weak to uniquely determine full-body motions~\cite{ao2022rhythmic,yoon2020speech}. Consider the following illustrative examples: 1) For generating co-speech motion of a digital host for releasing a new product, both "talking while walking" and "talking while standing still" are reasonable motions; 2) In a game, we want to produce scenes that NPCs are sitting, drinking tea, and talking with others. The target motion is full-body while cannot be identified by speech only; 3) A conversation avatar in a virtual environment is required to respond "welcome" to a new comer. For the similar reason, whether the avatar should keep standing or stepping forward, and waving hands or not, need to be controlled by additional prompts. As a result, it would be meaningful to realize precise and flexible control of full-body motion for achieving natural and synergistic effects based on additional input signals to reflect user intentions, such as text prompts. 

On the other hand, prompt-based co-speech motion generation is a highly nontrivial task with two major reasons. On one hand, the existing speech-to-motion datasets, such as BEATX~\cite{liu2024emage}, focuses on subtle hand movements yet involve fairly limited full-body motions, especially in lower body. For example, the lower body of the speaker usually remains relatively stable during talking. This makes a wide range of common human activities out of training distribution. On the other hand, these datasets also lack annotated user prompts. Furthermore, crafting of a diverse and annotated dataset at scale is extremely costly. This has significantly constrained the potential for quality and diversity in motion generation.

One possible solution to deal with the data lacking issue is to augment training with text-to-motion datasets, such as AMASS~\cite{AMASS:ICCV:2019}, which include a relatively complete set of full-body motions with vast scale and strong diversity, as well as annotated text prompts~\cite{BABEL:CVPR:2021,Guo_2022_CVPR}. Superficially, jointly training with both speech-to-motion and text-to-motion datasets could lead to the ideal model, whose key is to build a joint embedding space of speech, text, and motion. However, due to the significant distribution mismatch in motion between the two kinds of datasets, a large number of full-body motions are missing their corresponding speech signals, making building such an embedding space still a challenging task.

To deal with issue, we propose \ourapproach, a prompt-based co-speech motion generation approach which utilizes off-the-shelf text-to-motion datasets to augment co-speech training, meanwhile addressing the distributional mismatch challenge.
For training, we propose a multi-stage approach, which utilizes motion representation pre-training and motion-prompt alignment pre-training to address the issue of motion distribution mismatch and the problem of lacking prompt annotation for speech-to-motion data.  

For inference, we designed a novel separate-then-combine strategy under for both input conditions and body parts, such that the separate operations map the input signal to their most proper body part to control, meanwhile the combine operations leads to the synergy among body parts.
Extensive experiments show that, our approach is able to achieve significant performance in using both speech and text prompt to guide the generation of synergistic full-body motion precisely and flexibly, which is beyond the capability of the existing co-speech generation approaches.

In summary, by proposing \ourapproach, our main contributions are: 1) We propose the first approach to enable synergistic full-body control with general text prompts for co-speech motion generation, under the situation of lacking fully annotated datasets of speech, text, and motion; 
2) We propose a novel multi-stage training approach to address the motion distributional mismatch and prompt annotation lacking challenges; 
3) We propose a novel separate-then-combine approach for model inference to achieve both precise control and synergistic motion generation. 

The rest of the paper is organized as follows. In Section 2, we discuss the related work from two closest research areas, i.e. co-speech motion generation and text-to-motion generation. In Section 3-5, we introduce the model design, training process, and inference process of our approach in detail. In Section 6, we report experimental results. In section 7-8, we discuss limitations and future work as well concluding the paper. 

\section{Related Work}
\subsection{Co-Speech Motion Generation}

Early rule-based approaches to co-speech motion generation \cite{cassell2001beat, cassell1994rulefullbody, kopp2006bml} utilize linguistic rules to translate speech into sequences of predefined gesture segments. This process, being time-consuming and labor-intensive, requires significant manual effort in defining rules and segmenting motions. Previous generative methods often produce overly smooth motions \cite{ghorbani2022zeroeggs,liu2022beat,Speech2affectivegestures,tang2018gesturegan}, attributable to the use of traditional deterministic generative models, which are inadequate for many-to-many mapping problems. Despite some attempts to introduce control signals and prior information into model design \cite{ao2022rhythmic,liu2022beat,kucherenko2021speech2properties2gestures,yang2023QPGesture}, the capabilities of these models remain limited. Recent advancements have leveraged modern generative models like Diffusion \cite{dhariwal2021diffusion} to tackle these challenges. For instance, DiffGesture \cite{zhu2023taming} employs a diffusion model to capture the relationship between speech and gesture. Nonetheless, the weak semantic signals in audio often result in motions that are misaligned with the semantic content of the input audio. DiffuseStyleGesture \cite{yang2023diffusestylegesture} advances this by integrating emotional control into the gesture generation process, while Amuse \cite{chhatre2023emotional} and EMOTE \cite{EMOTE} explicitly extract and disentangle emotions from given conditions to provide stronger control signals. UnifiedGesture \cite{unifygesture} additionally use reinforce learning to strength gesture. GestureDiffuCLIP \cite{Ao2023GestureDiffuCLIP} incorporates existing contrastive learning frameworks \cite{tevet2022motionclip} to enable prompt-based gesture style control, offering finer-grained style controllability for end-users. However, these methods still struggle to meet diverse real-world user requirements, such as accommodating gestures while walking, due to the limited motion distribution in co-speech datasets.

\subsection{Text-to-Motion Generation}
In parallel to co-speech motion generation problems, text-based motion generation aims to generate general motions from textual prompts. Pioneering works~\cite{zhang2022motiondiffuse, tevet2023human, humantomato, chen2023humanmac,zhang2023generating} such as Motion Diffuse and T2M-GPT utilize a diffusion-based architecture or GPT-based architechture to model the many-to-many challenges in text-to-motion generation. Subsequent studies, such as PriorMDM, TLControl, and OmniControl~\cite{shafir2023human, wan2023tlcontrol, xie2023omnicontrol}, further employ trajectory and end-effector tracking to provide finer-grained control. GMD~\cite{karunratanakul2023gmd} introduces additional scene information during the generation of human actions, and MotionClip~\cite{tevet2022motionclip} attempts to align motions with the CLIP space~\cite{radford2021learning}, enabling the capability to generate motion from images. TM2D~\cite{Gong_2023_ICCV} and FreeTalker~\cite{yang2024Freetalker} have explored this by learning both speech-to-motion and text-to-motion tasks simultaneously. Even though this enables a single model to switch between two tasks, it does not provide synergistic generation conditioned on both signals.

\section{Model Design}
In this section, we introduce our prompt-based co-speech motion generation model. We first provide an overview of the model design and the corresponding generation process. Afterwards, we provide detailed descriptions of two core modules for motion representation and conditional generation.

\subsection{Overview}

Our model takes speech audio and the corresponding transcripts as inputs, targeting at outputting realistic and stylized full-body motions that align with the speech content rhythmically and semantically. Compared with traditional co-speech generation model, besides speech, it further allows to use a short piece of text, namely a \emph{text prompt}
to provide additional descriptions for the desired motion style. The full-body motions are then generated to follow the style given by both speech and prompt as much as possible.

\begin{figure}[t]
  \centering
  \includegraphics[width = .8\linewidth]{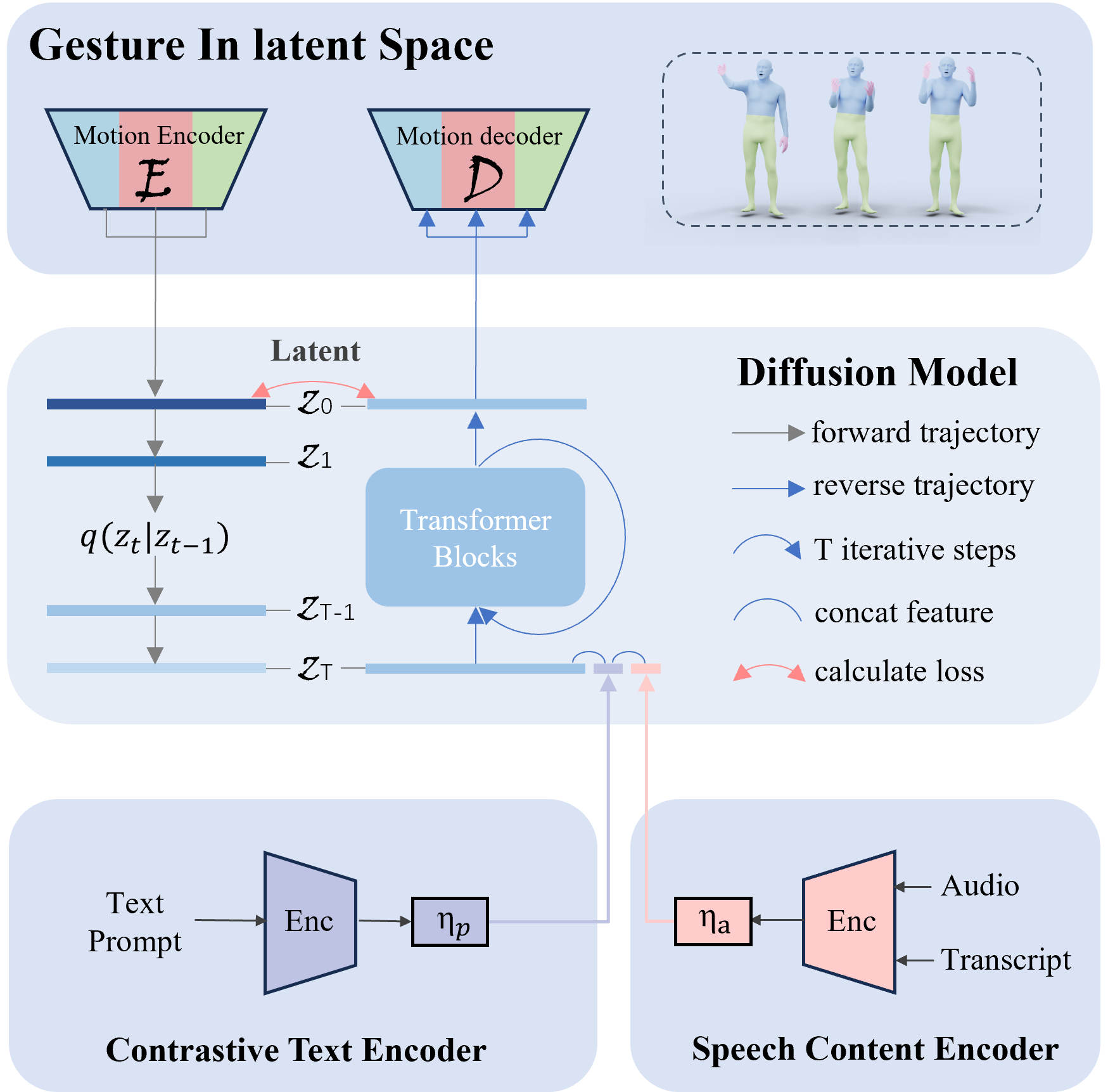}
  \caption{The structure of our prompt-based co-speech generation model.}
  \label{fig:model}
\end{figure}

The overall model structure is illustrated in Figure~\ref{fig:model}, which consists of three major components. The first is the {\it motion representation module}, which consists of motion encoder and decoders. We include three separate encoder and decoders to represent local body parts. The second is the {\it conditional generation module} for aligning latent motion representations with conditional inputs of speeches and prompts. The module is based on the latent diffusion model~\cite{rombach2021highresolution}, which apply diffusion and denoising steps in the latent space. The third is the {\it conditional representation module}, which consists of the speech content encoder and contrastive text encoder to obtain scalar-valued prompt and speech conditions in the diffusion-based conditional generation model. Below we dive into detailed structures of the first two modules.

\subsection{Motion Representation Module}
{\bf Motion encoding}. Recent studies in motion generation have demonstrated that vector-quantized autoencoder (VQ-VAE)~\cite{van2017neural} possesses a remarkable capability for compressing motion information~\cite{zhang2023generating,Ao2023GestureDiffuCLIP,guo2023momask}. We also utilize vector quantization for motion encoding. Following~\cite{guo2023momask,ng2024audio2photoreal,zeghidour2021soundstream}, we use a residual VQ-VAE (RVQ-VAE) as the quantization layer. To further decrease the coupling between body parts, we segment the body into three parts: upper body, fingers, and lower body, like in~\cite{Ao2023GestureDiffuCLIP,liu2024emage}, and train a separate RVQ-VAE for each part. In details, the motion sequence $\mathcal{M}$ can be represented as $\mathbf{m}_{1:N}\in\mathbb{R}^{N\times  D}$, which is firstly encoded into a latent vector sequence $\mathbf{{z}}_{1:n}\in\mathbb{R}^{n\times d}$ with downsampling ratio of $n/N$ and latent dimension $d$, using 1D convolutional encoder $\mathrm{E}$;
The $\mathbf{{z}}_{1:n}\in\mathbb{R}^{n\times d}$ obtained through the encoder then enters the first quantization layer $\mathrm{Q}_1$,
each vector subsequently finds its nearest code entry in the layer's codebook $\mathbf{C_1}=\{\mathbf{c^1_k\}}_{k=1}^{K}\subset \mathbb{R}^d$ to get the first quantization code $\mathbf{\hat{z}^{1}_{1:n}}$, also we can calculate the quantization $\mathbf{residual}_{1:n} = \mathbf{\hat{z}^1_{1:n}} - \mathbf{{z}}_{1:n}$. 
The $\mathbf{residual}_{1:n}$ then enter the second quantization layer $\mathbf{Q}_2$ finds its nearest code entry in the layer's codebook $\mathbf{C_2}=\{\mathbf{c^2_k\}}_{k=1}^{K}\subset \mathbb{R}^d$ to get the second quantization code $\mathbf{\hat{z}^2_{1:n}}$. Accordingly, $\mathbf{\hat{z}^3_{1:n}}$,$\mathbf{\hat{z}^4_{1:n}}$... can be calculated in this manner. 
As the last step of motion encoding, we sum all quantization code together to get the final code $\hat{z} = \sum_{q=1}^Q \mathbf{\hat{z}}^v$. 

{\noindent\bf Motion decoding}. Similar to motion encoding, three separated decoders are introduced for generating corresponding motions for all body parts, which are 1D convolutional decoders. During training, the motion data is encoded with motion encoders and fused with speech and prompt conditions by the diffusion-based conditional generation module, and then passed through the decoders to get the reconstructed motions. During inference, the motion encoder is not utilized, the generated motion is obtained directly from the speech and prompt conditions with the diffusion module and motion decoders.

\subsection{Conditional Generation Module}
The conditional generation module is based on the latent diffusion model~\cite{rombach2021highresolution}, which is a variant of diffusion models that applies the forward and reverse diffusion processes in a pre-trained latent feature space. The \emph{diffusion process} is modeled as a Markov noising process. Starting from a latent gesture sequence ${Z}_0$ drawn from the gesture dataset, the diffusion process progressively adds Gaussian noise to the real data until its distribution approximates $\mathcal{N}({0}, {I})$. The distribution of the latent sequences thus evolves as
\begin{align}
    q({Z}_n | {Z}_{n-1}) = \mathcal{N}(\sqrt{\alpha_n}{Z}_{n-1}, (1-\alpha_n){I}),
\end{align}
where ${Z}_n$ is the latent sequence sampled at diffusion step $n$, $n\in\{1, \dots, N\}$, and $\alpha_n$ is determined by the variance schedules. In contrast, the \emph{reverse diffusion process}, or the \emph{denoising process}, estimates the added noise in a noisy latent sequence. Starting from a sequence of random latent codes ${Z}_N \sim \mathcal{N}({0}, {I})$, the denoising process progressively removes the noise and recovers the original latent code ${Z}_0$. To achieve conditional motion generation, we train a network ${E}_{\theta}({Z}_n, n, {A}, {P})$, 
the \emph{denoising network}, to recover the noise-free codes based on the noisy latent motion codes $Z_n$, the diffusion step $n$, the audio $A$, and the prompt feature $P$ from the joint align space. Finally, the recovered code is input into the motion decoders for motion generation.

\section{Model Training}
\begin{figure*}[t]
  \centering
  \includegraphics[width=.85\linewidth]{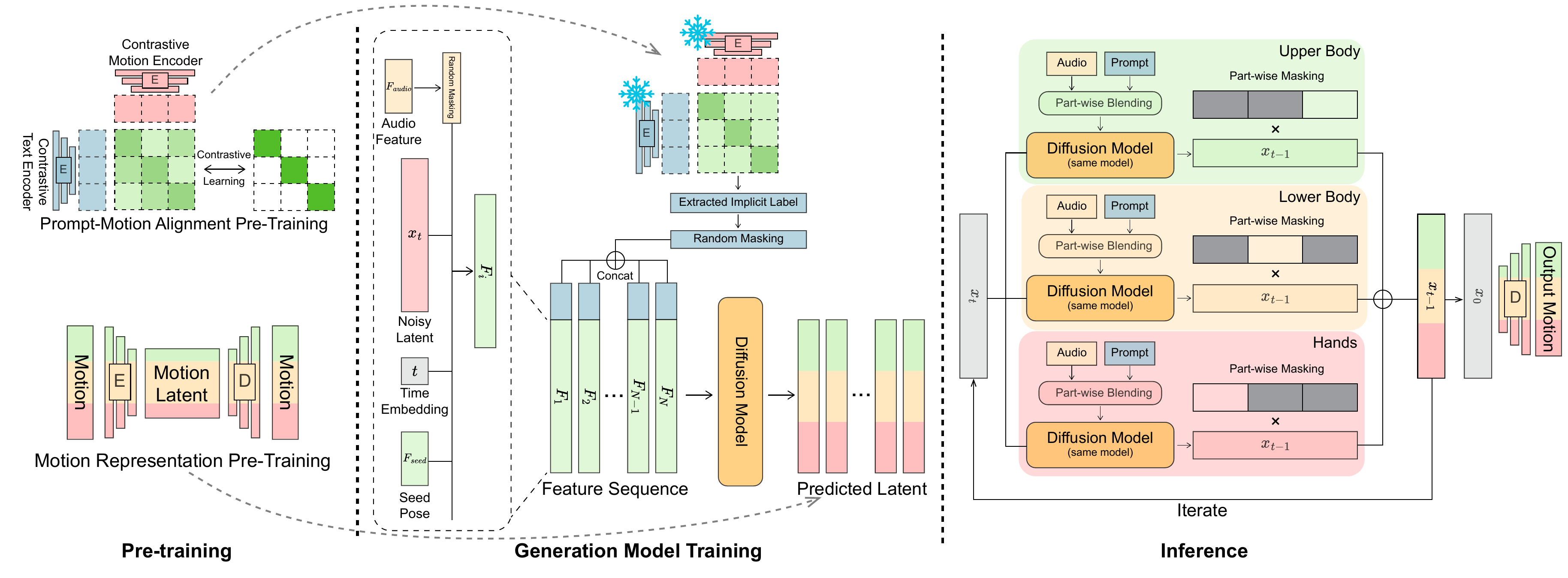}
  \caption{
    Illustration of the training and inference processes. We initially train a contrastive learning space between text and motion, alongside a motion auto-encoder that uses motion from both speech-to-motion and text-to-motion dataset for an expressive latent space. Subsequently, our co-speech latent diffusion model is trained under the guidance of an \textit{implicit label} extracted from motion using the contrastive space, effectively bypassing the lack of textual motion annotations in co-speech data. During inference, we implement a separate-then-combine strategy in every diffusion step, enabling finer control over individual body parts while preserving their synergistic interaction.
  }
  \label{fig:training}
\end{figure*}
The overall training pipeline is shown in Figure~\ref{fig:training}, which consists of a pre-training stage and a generation model training stage. The pre-training stage involves two tasks. The first task, motion representation pre-training, targets at training the motion encoders and decoders for all body parts based on motion data from 
both the speech-to-motion and text-to-motion datasets in order to address the issue of motion distribution mismatch.
The second task, prompt-motion alignment pre-training, targets at obtaining prompt-motion aligned embedding space~\cite{tevet2022motionclip,petrovich23tmr} based on the text-to-motion dataset and address the issue of lacking prompt annotations for speech-to-motion data. 
In the generation model training stage, both the speech-to-motion and text-to-motion data are utilized jointly with the motion encoders and decoders as well as the prompt-motion alignment space to obtain the final diffusion-based generation model~\cite{tevet2023human}. 
Below we discuss the details of the training process.

\subsection{Motion Representation Pre-Training}
The target of motion representation pre-training is to obtain the motion encoders and decoders based on motion data only, which are from the speech-to-motion and text-to-motion datasets. By this way, the obtained primitive motion representation space are independent of any conditional speech and prompt signals. From extensive experiments, we find that this approach effectively alleviates the motion distribution mismatch issue between speech-to-motion and text-to-motion datasets, which can not be addressed when directly mixing the two datasets for conditional generation model training without such a joint pre-training process.

Concretely, utilizing all motion data from the two kinds of datasets, the motion encoders and decoders are trained via a motion reconstruction loss combined with a latent embedding loss at each quantization layer of the RVQ-VAE structures:
\begin{align}
    \mathcal{L}_{rvq} = \|\mathbf{Z}-\mathbf{\hat{Z}}\|_1 + \beta\sum_{q=1}^Q\|\mathbf{z}^{q}-\mathrm{sg}[\mathbf{\hat{z}}^q]\|_2^2,
\end{align}
where $\mathrm{sg}[\cdot]$ denotes the stop-gradient operation, and $\beta$ a weighting factor for embedding constraint. This framework is optimized with straight-though gradient estimator~\cite{oord2018neural}, and our codebooks are updated via exponential moving average and codebook reset following T2M-GPT~\cite{zhang2023generating}. 
After training, the motion encoder and decoders are frozen in the rest of the training process.

\subsection{Text-Motion Alignment Pre-Training}

The target of prompt-motion alignment pre-training is to obtain the prompt-motion alignment embedding space, which consists of the contrastive text encoder in Figure~\ref{fig:model} and an additional contrastive motion encoder. These two encoders play the essential role to address the issue of missing prompt annotations for speech-to-motion data by employing the {\it implicit label} strategy. During downstream training, assume that the prompt annotation is needed for some speech-to-motion instance, which is lacking in the original dataset. We can directly input the motion into the contrastive motion encoder and get its corresponding embedding in the prompt-motion aligned space. It is easy to see that this motion embedding is an ideal substitution of the missing prompt embedding if the aligned space is well-trained. 

Concretely, motivated by~\cite{petrovich23tmr}, we formulate this pre-training task as a contrastive learning problem. 
Besides the contrastive text and motion encoders, we employ an additional motion decoder, which is different from the motion encoder in our final inference model. 
On the premise that the latent space is a probabilistic space, this setup aims to bring the feature vectors of corresponding text and motion pairs as close as possible. The decoder then decodes these latent feature vectors into motion to calculate the reconstruction loss with real motions. The loss gradients are back-propagated to update the prompt and motion encoders. This technique has been proven to be highly effective in numerous studies~\cite{petrovich22temos,tevet2022motionclip,petrovich23tmr}.

\textbf{Loss function design.} We introduce the same set of sub-loss terms to~\cite{petrovich23tmr}, and the total loss can be defined as the weighted sum formulation $\mathcal{L}_{\text{CON}} = \mathcal{L}_{\text{R}} + \lambda_{\text{KL}}\mathcal{L}_{\text{KL}} +\lambda_{\text{E}}\mathcal{L}_{\text{E}} + \lambda_{\text{NCE}} \mathcal{L}_{\text{NCE}}$. For sub-losses, the reconstruction loss $\mathcal{L}_{\text{R}}$ measures the motion reconstruction given prompt or motion input (via a smooth L1 loss). The Kullback-Leibler (KL) divergence loss $\mathcal{L}_{\text{KL}}$ is to regularize the distances between motion and prompt embedding distributions as well as making them closer to the standard normal distribution. 
The cross-modal embedding similarity loss $\mathcal{L}_{\text{E}}$ enforces both prompt $z^T$ and motion $z^M$ latent codes to be similar to each other (with a smooth L1 loss). A contrastive loss term $\mathcal{L}_{\text{NCE}}$ additionally uses negatives prompt-motion pairs to ensure a better structure of the latent space. More detailed introductions of these loss terms are included in the appendix.

After training, the contrastive text and motion encoders are frozen and utilized in the downstream generation model training.

\subsection{Generation Model Training}
After the pre-training stage, we obtain two sets of outcomes: the motion encoders and decoders, as well as the contrastive text and motion encoders. Based on the motion encoders obtained from motion representation pre-training, we can map all motions in the global motion distribution to the same compact latent space. Utilizing the contrastive text and motion encoders from the prompt-motion alignment pre-training, for motions without prompt annotations in the speech-to-motion dataset, we can provide them with an implicit label using the contrastive motion encoder. What is essential here is that 1) the motion distribution mismatch problem is addressed for motion representations; 2) all co-speech training data have their corresponding (implicit) prompt annotations. 

The training of the generation model mostly follows the standard training process of denoising diffusion models~\cite{dhariwal2021diffusion,rombach2021highresolution}. We train the denoising network ${E}_{\theta}$ by drawing random tuples $({Z}_0,n,{A},{P})$ from the training dataset, corrupting ${Z}_0$ into ${Z}_n$ by adding random Gaussian noises ${E}$ to obtain ${Z}_n$, applying denoising steps to ${Z}_n$ using ${E}_{\theta}$, and optimizing the loss 
\begin{equation}
    \mathcal{L}_{{net}} = 
    L1_{smooth}[{Z}_0 - {E}_{\theta}({Z}_n, n, {A}, {P})].
\end{equation}
Specifically, the latent motion representation ${Z}_0$ is encoded by the motion encoder with the RVQ-VAE structure, and the prompt embedding is obtained from the implicit labels generated by the contrastive motion encoder. Since the speech audio and speech text transcript always occur simultaneously during speech, we uniformly denote them as $A$ here. $A$ is processed through a temporal convolutional network for feature extraction and to align with the latent motion sequences in the time series.

We utilize the classifier-free guidance~\cite{ho2022classifierfree} to train our model. To strengthen the understanding of the two conditional signals, speech $A$ and prompt $P$, we make the diffusion model to learn under both conditioned and unconditioned distributions during training by randomly setting conditional variables $\bf{A}$ and $\bf{P} = {\mathbf 0}$ for $\eta_a$ and $\eta_p$.
This makes the diffusion model better understand the impact of various conditional signals on the generation results. More details of the training techniques are introduced in the appendix.

\section{Model Inference}
Through the multi-stage training process, the generation model is obtained. However, utilizing this for conditional generation is not a straightforward task. Even though an aligned space of motion, speech, and prompt is obtained, precise control and generation still requires carefully aligning generation conditions to local body parts. To achieve this target, we introduce the separate-then-combine generation strategy for manipulating latent codes of the diffusion model for both input conditions and body parts.

{\bf General diffusion-based generation process.} During inference, the diffusion network leverages the sampling algorithm of DDPM~\cite{ho2020denoising} to synthesize motions. It first draws a sequence of random latent codes ${Z}_N^*\sim{}\mathcal{N}({0}, {I})$ then computes a series of denoised sequences $\{{Z}_n^*\},{n=N-1,\dots,0}$ by iteratively removing the estimated noise ${E}_n^*$ from ${Z}_n^*$. The entire process is carried out in an autoregressive manner.

Sampling from $p(Z_{0}|n,A,P)$ is done in an iterative manner, according to ~\cite{ho2020denoising}. In every time step $n$ we predict the clean sample $\hat{Z}_{0} = G(Z_t, n, A, P)$ and noise it back to $Z_{t-1}$. This is repeated from $t=N$ until $Z_0$ is achieved.

{\noindent\bf Separate-then-combine for conditions.} Motivated by MotionDiffuse~\cite{zhang2022motiondiffuse} and PIDM~\cite{bhunia2022pidm}, we extend our system to allow separated guidance to apply the effect of the conditional signal audio and prompt. To achieve this, from the dimension of conditions, we separate latent codes into the following formulation: 
\begin{equation}
    \bf{Z}_{\text{cond}} = \bf{Z}_{\text{uncond}} + w_a\bf{Z}_{\text{speech}} + w_p\bf{Z}_{\text{prompt}},
    \label{eq:sample1}
\end{equation}
where $Z_{\text{uncond}}=Z_{\theta}({Z}_n,n,{\mathbf 0},{\mathbf 0})$ is the unconditioned prediction of the model, such that both the speech and prompt conditions are set as the all-zero tensor $\mathbf 0$. The audio-guided prediction and the prompt-guided prediction are respectively represented by $Z_{\text{speech}}=Z_{\theta}(\bf{Z}_t,t,\bf{A},{\mathbf 0})-Z_{\text{uncond}}$ and $Z_{\text{prompt}}=Z_{\theta}(\bf{Z}_t,t,{\mathbf 0},\bf{P})-Z_{\text{uncond}}$. $w_a$ and $w_p$ are guidance scale corresponding to speech and prompt. 

{\noindent\bf Separate-then-combine for body parts.} Furthermore, we extend our system to allow fine-grained style control on individual body parts.
We utilize the diffusion model to generate codes for each body part based on masking. The full-body motion codes $\bf{Z}^{\mathcal{O}} \in \mathbb{R}^{O \times (L \times C)}$ is then computed by stacking the motion codes of each body part. At inference time, we predict full-body signal $\{\bf{E}_{cond,o}^*\}_{o\in\mathcal{O}}$ conditioned on a set of style prompts $\{\bf{P}_o\}_{o\in\mathcal{O}}$ for every body part, where each $\bf{E}_{cond,o}^*$ is calculated by Equation\eqref{eq:sample1}. These body part signals can be simply fused as $\bf{E}_{cond}^* = \sum_{o\in \mathcal{O}}\bf{E}_{cond,o}^* \cdot M_o$, where $\{M_o\}_{o\in\mathcal{O}}$ are binary masks indicating the partition of bodies in $\mathcal{O}$. To achieve better motion quality, we add a smoothness item to the denoising direction as suggested by~\cite{zhang2022motiondiffuse},
\begin{equation}
    \bf{Z}_{cond}^* = \sum_{o\in\mathcal{O}}\Bigl(\bf{Z}_{cond,o}^* \cdot M_o\Bigr) 
\end{equation}

\begin{figure*}[t]
    \centering
    \includegraphics[width=.85\linewidth]{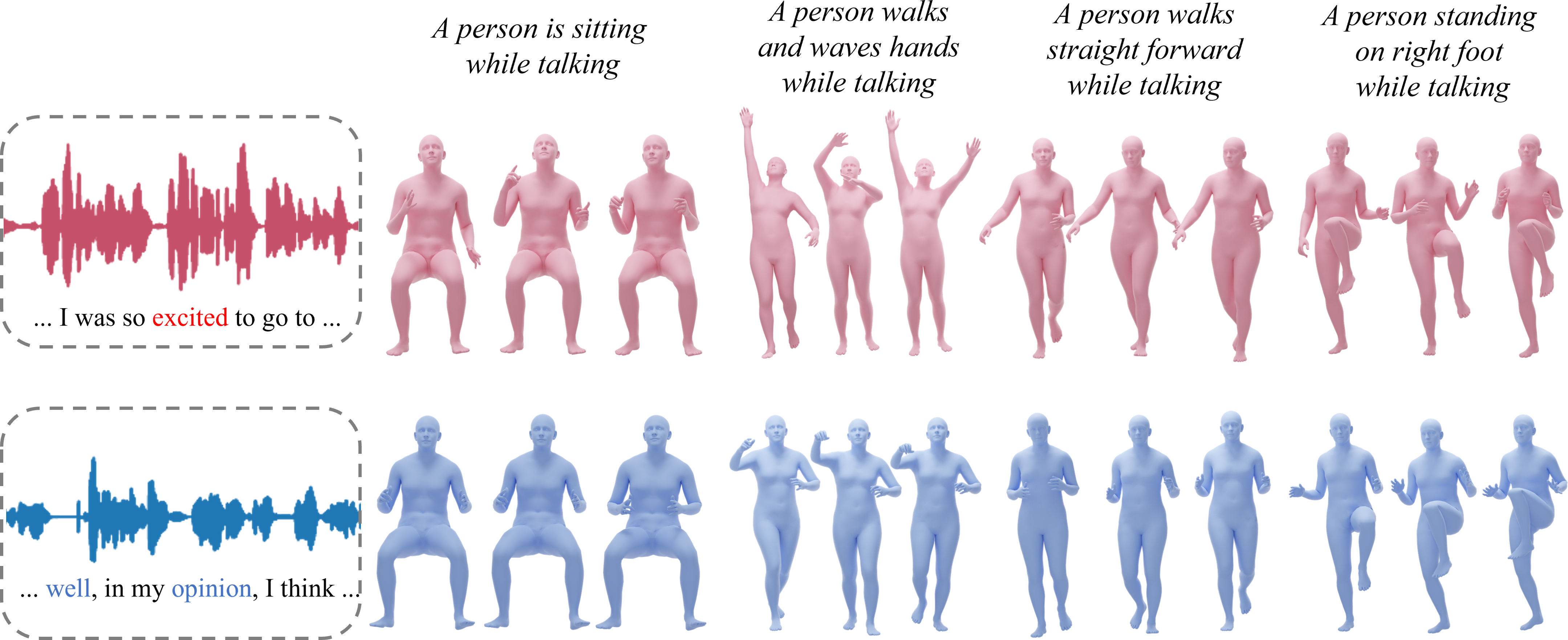}
    \caption{Qualitative results for synergistic full-body motion generation. More results are included in the appendix as well as demo videos. 
    }
    \label{fig:demo_show}
\end{figure*}

Afterwards, the following generation procedure follows the normal diffusion generation process as discussed above. By utilizing the two separate-then-combine strategies, the control of the motion generation process can be more flexible. The separation steps effectively build more precise mapping between input conditions and body parts. On the other hand, the combine steps ensure to generate synergistic full-body motions, avoiding unnatural motions to appear in generation.

\section{Experiments}

In this section, we report experimental results to verify the following questions:
\begin{itemize}
\item
{\bf Full-body synergistic generation.} Is \ourapproach\;able to generate full-body synergistic motion desirably aligned with both speech and prompt inputs, which is the first co-speech motion generation approach to achieve this functionality?
\item 
{\bf Sinlge-source conditional generation.} Is \ourapproach\;able to achieve comparable or even better performance over state-of-the-art approaches under single-source conditional generation, verifying that besides motions, \ourapproach\;truly learns desirable representations for both speeches and prompts?
\item
{\bf Ablation study.} Is the multi-stage training strategy and the separate-then-combine inference strategy indeed essential to achieve desirable performance?
\end{itemize}
Below we explore all three questions in details. More experimental results can also be found in the appendix. 
\subsection{Experimental Setup}
\noindent\textbf{Dataset}. Our model smartly avoids the need for annotated co-speech motion data by leveraging existing speech-to-motion and text-to-motion datasets. For the speech-to-motion dataset, we utilize BEATX-Standard \cite{liu2024emage}, which includes 30 hours of co-speech motion, paired with audio and transcripts. For the text-to-motion dataset, we employ HumanML3D \cite{Guo_2022_CVPR}, which annotates motion in the AMASS dataset \cite{AMASS:ICCV:2019}, consisting of 14,616 annotated motion sequences and 44,970 annotations. All motions are in SMPLX format \cite{SMPL-X:2019} and consist of 30 frames per second.

\noindent\textbf {Implementation Details}. We utilize the RVQVAE~\cite{zeghidour2021soundstream} as our auto-encoder architecture, featuring resblocks in both the encoder and decoder with a downscale factor of 4. Residual quantization employs 6 quantization layers, each with a code dimension of 512 and a codebook size of 512, with a quantization dropout ratio set at 0.2. During contrastive pre-training, we establish a space with a dimension size of 256 and a batch size of 32. We also set the temperature $\tau$ to 0.1, contrastive loss weight to 0.1, and negative-filtering threshold to 0.8.
Our diffusion model incorporates 8 transformer layers and is trained with a batch size of 200 and a latent dimension of 512. The number of diffusion steps is 1000. All components can be trained on a single 4090 GPU within three days.

\subsection{Full-Body Synergistic Generation}
In the first experiment, we aim to verify whether our approach effectively supports the generation of synergistic full-body motions conditioned on both speech and flexibly-chosen prompts. As no previous research has addressed this specific task, we focus on evaluating the generation results of our approach using carefully designed input speeches and prompts. It's important to note that we also provide experimental analysis on GestureDiffuseCLIP~\cite{Ao2023GestureDiffuCLIP} and FreeTalker~\cite{yang2024Freetalker} in the appendix. These works address related but distinct tasks. GestureDiffuseCLIP also supports prompt-based co-speech motion control. However, this control is limited to motions {\it inside of the speech-to-motion dataset}. In comparison, our approach supports general out-of-distribution motions. FreeTalker, on the other hand, trained a model capable of switching between speech-to-motion and text-to-motion generation tasks, but it does not produce synergistic results under both speech and prompt conditions.

\begin{table*}[t]
    \centering
    \caption{\textbf{Comparison with the state-of-the-art methods on HumanML3D~\cite{Guo_2022_CVPR} test set.} We compute standard metrics following~\cite{Guo_2022_CVPR}. For each metric, we repeat the evaluation 20 times and report the average with 95\% confidence interval. For MDM and MLD, we report the results using ground-truth motion length.}

    {
    \begin{tabular}{l c c c c c c c}
    \toprule
    \multirow{2}{*}{Methods}  & \multicolumn{3}{c}{R-Precision $\uparrow$} & \multirow{2}{*}{FID $\downarrow$} & \multirow{2}{*}{MM-Dist $\downarrow$} & \multirow{2}{*}{Diversity $\uparrow$}\\

    \cline{2-4}
    ~ & Top-1 & Top-2 & Top-3 \\
    
    \midrule
        Real motion & \et{0.490}{.003} & \et{0.682}{.003} & \et{0.783}{.003} & \et{0.001}{.001} & \et{3.378}{.007} & \et{10.471}{.083} \\

        MDM~\cite{tevet2023human} & \et{0.363}{.007} & \et{0.553}{.008} & \et{0.662}{.007} & \et{1.390}{.088} & \et{4.599}{.037} & \et{10.704}{.066}\\ 

        T2M-GPT ~\cite{zhang2023generating} & \et{0.433}{.003} & \et{0.615}{.002} & \et{0.716}{.003} & \et{0.564}{.012} & \et{3.867}{.008} & \et{10.558}{.083}\\ 

        MLD~\cite{chen2023executing} & \et{0.429}{.003} & \et{0.613}{.003} & \et{0.717}{.002} & \et{0.963}{.029} & \et{3.898}{.012} & \et{10.401}{.096}  \\

        MoMask ~\cite{guo2023momask} & \et{0.461}{.002} & \et{0.657}{.003} & \et{0.760}{.002} & \et{0.222}{.007} & \et{3.620}{.011} & \et{10.621}{.096} \\

    \midrule

        \ourapproach\;(w/o text-to-motion alignment) & \et{0.429}{.003} & \et{0.622}{.004} & \et{0.732}{.004} & \et{0.509}{.013} & \et{4.033}{.013} & \et{10.231}{.096}\\ 

        \ourapproach\;(w/o motion representation pre-training) & \et{0.097}{.002} & \et{0.178}{.002} & \et{0.253}{003} & \et{17.797}{.056} & \et{7.146}{.010} & \et{6.127}{.057}\\

        \ourapproach & \et{0.375}{.003} & \et{0.564}{.003} & \et{0.681}{.002} & \et{4.385}{.034} & \et{4.499}{.012} & \et{9.374}{.073}  \\

    \bottomrule

    \end{tabular}
    }
    \vspace{1mm}

    \label{tab:text}

\end{table*}
\begin{table}[t]
\centering
\caption{\textbf{Comparison with the state-of-the art methods on BEATX~\cite{liu2024emage} test set. Quantitative evaluation on BEATX.} We report FGD $\times 10^{-1}$, BC $\times 10^{-1}$, and diversity.}
{
\begin{tabular}{lccc}
\toprule
     Method & FGD $\downarrow$ & BC $\uparrow$ & Diversity~$\uparrow$   \\ 
\midrule
GT & 0.000 & 6.897 & 12.755 \\

recons & 1.729&7.122 &12.599\\
recons(w/o residual)  &3.913&6.758&13.145\\

\midrule
S2G\cite{ginosar2019learning} & 25.129  & 6.902  & 7.783                        \\
Trimodal\cite{yoon2020speech} & 19.759  & 6.442  & 8.894                         \\
HA2G\cite{ha2g:liu2022learning} & 19.364  & 6.601 & 9.671                        \\
DisCo\cite{liu2022disco} & 21.170  & 6.571 & 10.378                              \\
CaMN\cite{liu2022beat} & 8.752  & 6.731 & 9.279                                  \\
DiffStyleGesture\cite{yang2023diffusestylegesture} & 10.137 & 6.891 & 11.075     \\
Habibie \textit{et al}.\cite{habibie2021learning} & 14.581 & 6.779 & 8.874      \\
TalkShow\cite{talkshow:yi2022generating} & 7.313 & 6.783 & 12.859         
\\       
EMAGE \cite{liu2024emage} & 5.423 & 6.794 & 13.057                         
\\ 
\midrule

\ourapproach\;(w/o mo.rep.) & 5.759 &7.181  &10.731 \\
\ourapproach\;(w/o align.) & 5.242 &\textbf{8.010}  &\textbf{13.521} \\
\ourapproach\;(w/o both) & \textbf{4.687}  &{7.363}  & {12.425}\\  
\ourapproach\;& 6.413 &7.971  &12.721 \\
\bottomrule
\end{tabular}}
\label{tab:audio}
\end{table}

As shown in Figure \ref{fig:demo_show}, we conduct qualitative experiments to evaluate the synergistic generation results of our model. To better demonstrate that the generation results synergistically integrate both speech and text prompt guidance, we present outcomes under two distinct speech audios: one \textit{excited} and the other \textit{calm}. We evaluate our method using four different text prompts: sitting, waving while walking, standing on the right foot, and walking straight forward. The results show that our model produces talking motions that closely align with the input speech audio while accurately adhering to the text prompt requirements for body gestures. With the excited audio, the motions exhibit more pronounced changes compared to the calm speech. These include increased arm movements, higher arm raises, more pronounced left-right body turns with larger arm movements, and a tendency for hands to reach outward while talking. For additional results, please refer to the appendix.

\subsection{Single-Source Conditional Generation}
In the second experiment, we focus on verifying whether our approach indeed learns a desirable joint embedding space, in special for the input conditions. To achieve this purpose, we introduce two single-source conditional generation benchmarks, i.e. speech-to-motion generation without prompts and text-to-motion generation without speeches. By quantitative comparison with state-of-the-art approaches under these two distinguished domains, we are able to verify whether our approach successfully distills information from both speech and prompts, meanwhile avoiding interference among them, which would be revealed by performance degeneration in single-condition generation. Note that for our approach, single-source generation is  realized by the similar method utilized in generating $\bf{Z}_{\text{speech}}$ and $\bf{Z}_{\text{prompt}}$ in Equation~\ref{eq:sample1}. The implementation details of all contenders are included in the appendix.

{\noindent\bf Speech-to-motion.}
We compare our approach with state-of-the-art speech-to-motion generation approaches, whose results are cited from~\cite{liu2024emage}. As shown in Table~\ref{tab:audio}, our method significantly outperforms baselines in terms of FGD~\cite{yoon2020speech}, BC~\cite{li2021aist++}, and diversity~\cite{li2021audio2gestures}. This result provides convincing proof that our approach generates significantly desirable speech representation to support strong speech-to-motion generation. To further verify how speech representations are affected by the multi-stage training process, we also conducted ablation studies under this task. The details are discussed in Section 6.5. 

{\noindent\bf Text-to-motion.} In this task, we compare our method with four state-of-the-art text-to-motion generation approaches. The results are reported in Table 1. It can be observed that our approach could achieve comparable performance to the existing baselines, showing its effectiveness in understanding text prompts. Similar to speech-to-motion, we conduct ablation studies for further justification, whose results are reported in the next subsection.

\subsection{Ablation Study}

\begin{figure}[t]
    \centering
    \includegraphics[width=.8\linewidth]{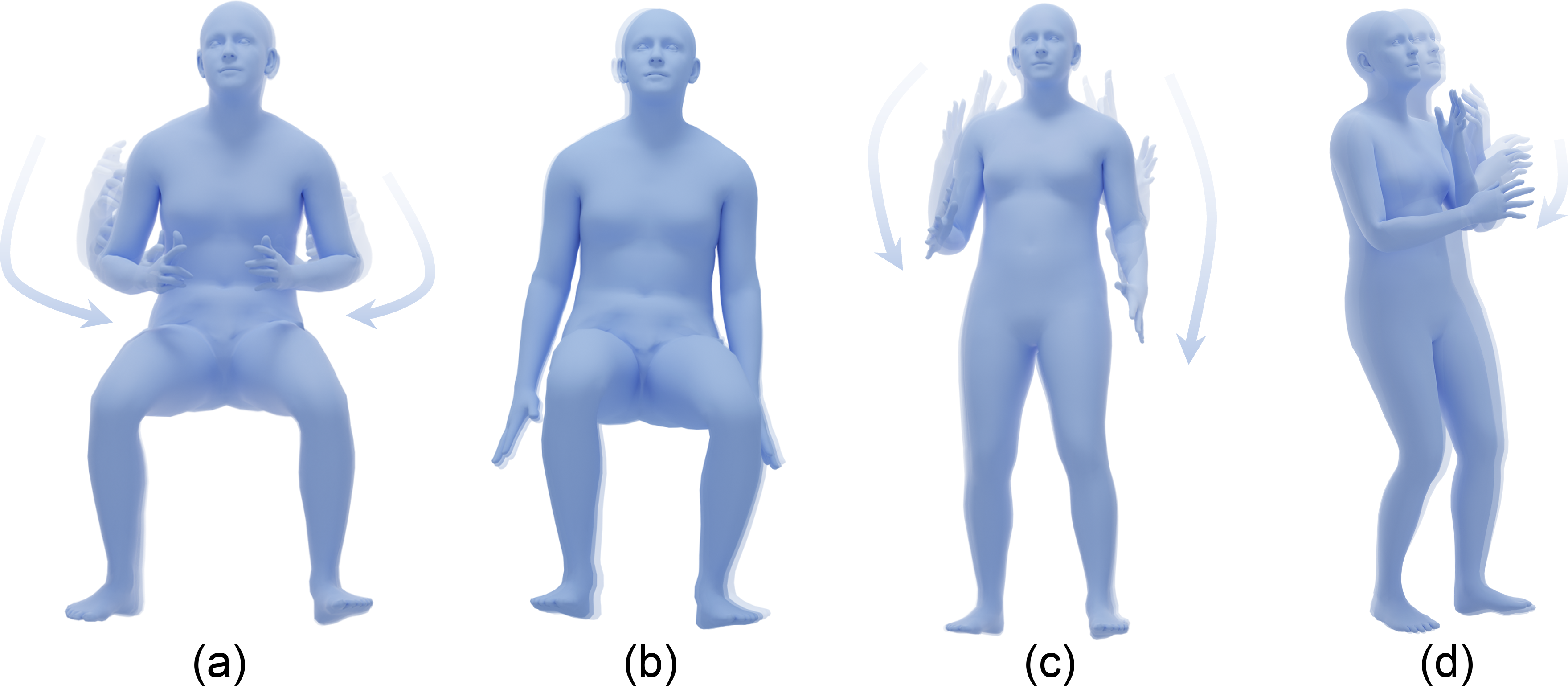}
    \caption{Qualitative ablation studies on training and inference procedures. More results are included in the appendix.}
    \label{fig:ablation1}
\end{figure}

In this section, we demonstrate qualitative examples of ablation study on model components and assess their contribution to the synergistic generation capability. For clarity, we demonstrate the results using a single prompt. Please refer to the appendix for additional results. As shown in the Figure ~\ref{fig:ablation1}, given the same speech and the text prompt "a person is talking while sitting", compared to the sitting and talking motion in Figure 5(a), removing our proposed components result in non-ideal generation results. Figure 5(b) corresponds to the removal of \textit{implicit labeling} in train stage, Figure 5(c) corresponds to the removal of \textit{separate-then-combine strategy} in inference stage, and Figure 5(d) corresponds to the removal of \textit{motion representation pre-training} in pre-training stage.

\noindent \textbf{Implicit labeling.} Figure 5(b) demonstrates the impact of removing implicit labeling during the training stage. Without implicit labeling, the model defaults to merely reacting to the textual prompt, producing only a static sitting motion. This confirms that without our proposed implicit labeling method, the diffusion model does not automatically learn to synthesize and integrate input to produce synergistic motions conditioned on both signals.

\noindent \textbf{Separate-then-combine strategy.} Figure 5(c) illustrates the effect of omitting the separate-then-combine strategy. Although the model learns to respond to both audio and prompt signals when trained with the implicit labeling method, the textual prompt inherently imposes different requirements on various body parts. The absence of the separate-then-combine strategy eliminates part-level guidance, leading the diffusion model to incorrectly merge multiple features. In this scenario, the model misinterprets the instruction to sit as merely lowering the arms and slightly bending the legs, rather than sitting.

\noindent \textbf{Motion representation pre-training.} Figure 5(d) shows the results when the joint training stage is omitted from the training process. The character shows an inclination to sit, but such motion is not represented within the limited distribution of the speech-to-motion datasets, rendering accurate generation unfeasible.

\noindent \textbf{Single-condition experiment ablation.}
Though our work focuses on the synergistic co-speech motion generation that follows audio and text motion prompts at the same time, we also evaluate the impact of our proposed methods on the single-source condition generation ability, whose results are shown in both Table 1 and 2.

We first remove all our propose components for synergistic co-speech motion generation, which results in a pure co-speech motion generation model conditioned on audio signal. This base model achieves the state-of-the-art performance in pure co-speech motion generation task, which serves as a solid base for our synergistic generation. Note that our base model only performs co-speech motion generation and it is unable to operate on text-to-motion task, resulting in one less ablation result in Table 1.

We then evaluate the result with prompt-motion alignment while motion representation pre-training is removed. This removal has a positive impact on the text-to-motion generation, as prompt-motion alignment enables the model to accept both the audio and text prompt as input, which drags the model's output distribution to co-speech generation instead of pure prompt-based motion generation.

Finally, we evaluate the single-condition performance when text-to-motion alignment is removed while motion representation pre-training is kept. This results in significant performance drop in text-to-motion generation due to the lack of solution space in text-to-motion generation. In co-speech motion generation, this results in a minor performance drop due to the solution space expanding beyond the original co-speech motion distribution, which is desirable in generating synergistic co-speech full body motion.

\section{Limitations and Future Work}

Although our model implements semantically correct body-level control through the separate-then-combine strategy, it primarily treats conditional inputs as signals of varying strength rather than fully comprehending them. This limitation highlights the inherent challenge in the field of motion generation: the difficulty of generative models in accurately interpreting multi-modal conditions. This difficulty underscores the need for future research to focus on the deeper comprehension of user prompts, which remains a formidable and crucial challenge. Additionally, hand gestures play a significant role in the realism and expressiveness of generated gestures. Currently, the lack of an annotated hand gesture dataset limits the ability to control gestures via textual prompts, which can see significant improvements through future related work.

Perhaps we can adopt certain appropriate methods for data augmentation, thereby further expanding the model's understanding capability when both speech and text prompts are input simultaneously. For example, we could apply the joint grouping and assembling strategy used in SINC~\cite{SINC:2023} for different motions to text-to-motion and speech-to-motion data. SINC focuses on data in the text-to-motion domain, emphasizing semantic consistency. However, if we involve speech-to-motion data, we also need to consider rhythm consistency. Finding a way to correctly determine whether the synthetic data aligns with both semantic and rhythm consistency for data augmentation is a promising area for future work.

\section{Conclusion}
In this paper, we propose \textit{\ourapproach}, targeting at addressing the lack of elaborate control issue of current co-speech motion generation approaches. Our main contributions are: 1) By introducing a multi-stage training process, we effectively utilize off-the-shelf text-to-motion datasets to enable the diffusion model to simultaneously understand both co-speech audio signals and textual requirements. This approach allows for the generation of synergistic full-body co-speech motions; 2) A separate-then-combine strategy during the inference stage, enabling fine-grained control over different local body parts. Extensive experiments demonstrate the effectiveness of our method and show that it can achieve precise control over the generated synergistic full-body motions, surpassing the capabilities of existing methods.

\begin{acks}
This work is supported by the National Key Research and Development Program of China (2022YFF0902302) and the National Natural Science Foundation of China (62322209,62206245).
We would like to thank Ling-Hao Chen for suggesting us using RVQVAE, and helping us training serveal versions of OpenTMA~\cite{chen2024opentma}, Haiyang Liu for helping us reproduce the baseline EMAGE, and Tenglong Ao for assisting us with the reproduction of GestureDiffuCLIP.
\end{acks}

\bibliographystyle{ACM-Reference-Format}
\bibliography{sample-base}


\clearpage

\appendix

\section{Implementation Details}

\subsection{Loss Function in Section 4.2}

\noindent\textbf{Contrastive Losses.} We adopted the foundational set of losses basing on~\cite{petrovich23tmr}, which are expressed as the weighted sum of 4 losses $\mathcal{L}_{\text{CON}} = \mathcal{L}_{\text{R}} + \lambda_{\text{KL}}\mathcal{L}_{\text{KL}} + \lambda_{\text{E}}\mathcal{L}_{\text{E}} + \lambda_{\text{NCE}} \mathcal{L}_{\text{NCE}}$.

More specifically,

\begin{equation} \label{eq:lr}
\mathcal{L}_{\text{R}} = \mathscr{L}_1(H_{1:F}, \widehat{H}^{M}_{1:F}) + \mathscr{L}_1(H_{1:F}, \widehat{H}^{T}_{1:F}) 
\end{equation}

\begin{equation}
\begin{split} \label{eq:kl}
    \mathcal{L}_{\text{KL}} &= \text{KL}(\phi^T, \phi^M) + \text{KL}(\phi^M, \phi^T) \\
    &+ \text{KL}(\phi^T, \psi) + \text{KL}(\phi^M, \psi).
\end{split}
\end{equation}

\begin{equation}
\begin{split} \label{eq:manifold}
    \mathcal{L}_{\text{E}} &= \mathscr{L}_1(z^T, z^M).
\end{split}
\end{equation}

\begin{equation}
\small
\begin{split} 
    \mathcal{L}_{\text{NCE}} &= - \frac{1}{2N} \sum_{i} \left( \log \frac{\exp{S_{ii}/\tau}}{\sum_{j} \exp{S_{ij}/\tau}} + \log \frac{\exp{S_{ii}/\tau}}{\sum_{j} \exp{S_{ji}/\tau}} \right) ~,
\end{split}
\label{eq:infonce}
\end{equation}
Reconstruction loss $\mathcal{L}_{\text{R}}$ quantifies the accuracy of motion reconstruction from text or motion inputs using a smooth L1 loss. The Kullback-Leibler (KL) divergence loss $\mathcal{L}_{\text{KL}}$ includes four components: two to regularize each encoded distribution—$\mathcal{N}_(\mu^M, \Sigma^M)$ for motion and $\mathcal{N}_(\mu^T, \Sigma^T)$ for text—to align with a standard normal distribution $\mathcal{N}_(0, I)$. The other two components enforce distributional similarity across the two modalities. A cross-modal embedding similarity loss $\mathcal{L}_{\text{E}}$ mandates that the latent codes for text $z^T$ and motion $z^M$ exhibit similarity (utilizing a smooth L1 loss). Additionally, a contrastive loss $\mathcal{L}_{\text{NCE}}$ leverages negative motion-text pairs to enhance the structuring of the latent space, where $\tau$ is the temperature hyperparameter and $S$ denotes for similarity.

The coefficients $\lambda_{\text{KL}}$ and $\lambda_{\text{E}}$ were set to $10^{-5}$, and $\lambda_{\text{NCE}}$ to $10^{-1}$ in our experiments, consistent with the settings in~\cite{petrovich23tmr}.

\subsection{Model Training Details in Section 4.3}

During this training stage, the pre-trained model components are frozen. To ensure feature consistency, we consistently use implicit labeling, regardless of whether the current ground-truth motion has a corresponding text label. Audio signals are set to zero in the absence of input, and are randomly masked during training to implement classifier-free guidance.

\subsection{Model Inference}

In the actual inference process, users may enter text prompts that specify control over multiple body parts, such as "a person is sitting and raising their left hand." It is challenging for the model to automatically determine which body parts need to be controlled while avoiding unnecessary manipulation of other parts, often resulting in awkward generative outcomes~\cite{SINC:2023}. To address this, in our separate-then-combine strategy, we use an effective method for part-wise control. We preprocess the prompt using a language model (T5X-Large), which deconstructs the prompt into multiple sub-prompts targeting individual body parts. Part-wise blending is then applied based on the presence of a sub-prompt for each body part; in the absence of a sub-prompt, the part-wise blending mechanism effectively zeroes out the text feature for that part. This approach also allows users to manually assign text prompts to specific body parts, enabling more precise and granular control. In addition, during the blending process of inference signals, similar to many works using classifier-free guidance~\cite{tevet2023human,yang2023diffusestylegesture}, our model supports a finer-grained control by manually manipulating the part-wise blending weights to control the intensity of the signal.

In evaluating single-modality signal metrics, we apply a single-source condition for fairness, omitting our proposed separate-then-combine strategy. For the quantitative assessment of speech-to-motion generation, we use the test set division from EMAGE~\cite{liu2024emage}, the speech-to-motion dataset employed. During this evaluation, part-wise blending is solely conditioned on the audio signal. For the quantitative assessment of text-to-motion results, we follow the test set division of HumanML3D~\cite{Guo_2022_CVPR}. Here, part-wise blending is solely conditioned on the text signal.

\subsection{Implementation of Baselines in Section 6.3}
{\bf Baselines in Table 1.}

For MDM~\cite{tevet2023human} and T2M-GPT~\cite{zhang2023generating}, due to the inherent seeding in our method, for a fair comparison during testing, with the MDM method, we infer 128 frames at once, incorporating the first 16 frames into the inference process via inpainting. For the T2M-GPT method, we treat the first four tokens as known and allow the model to autoregressively generate the subsequent tokens. 

\noindent{\bf Baselines in Table 2.}

We directly report their results shown in their original papers. Since the original results are without variances, we keep this tradition in the table.

\section{The choice of align space}
In our work, We additionally tested MotionCLIP~\cite{tevet2022motionclip} as our alignment space. Given that the CLIP space inherently aligns text and image features, using MotionCLIP as the alignment space introduces the extra modality of images to control motion generation.

However, the experiments demonstrated that MotionCLIP itself has the following issues: 1.Training instability: After training for more than 100 epochs, the generated motions tend to become still, a phenomenon also reported in HumanTOMATO~\cite{humantomato}.
2.Unsuitability for our work: Due to the instability in training MotionCLIP, we selected checkpoints to apply to our downstream tasks. These checkpoints included like the lowest CLIP Loss, the lowest total Loss, and checkpoints when the training loss had converged after long training periods. However, applying these MotionCLIP checkpoints to our tasks failed to achieve satisfactory results, as shown in the following Table~\ref{tab:text}. The metrics in pure text-to-motion tasks were also extremely low, which might be due to the inherent incompatibility of the CLIP space with the motion modality.

We additionally attempted to investigate why MotionCLIP is unsuitable for our downstream tasks. Since our downstream tasks only utilize the text encoder and motion encoder of MotionCLIP, we focused on analyzing the feature space distribution extracted by these encoders.

Using the text encoder from MotionCLIP, we randomly selected 100 texts from the dataset and calculated their pairwise cosine similarities. Among these 10,000 pairs, 96.24\% had cosine similarities greater than 0.70, and 99.38\% were greater than 0.65, with a mean distribution of 0.820 and a standard deviation of 0.056. Correspondingly, using the motion encoder from MotionCLIP, we randomly selected 100 motion clips from the dataset and calculated their pairwise cosine similarities. Among these 10,000 pairs, 84.04\% had cosine similarities greater than 0.95, and 99.96\% were greater than 0.90, with a mean distribution of 0.963 and a standard deviation of 0.015.

From the above analysis, it can be concluded that the CLIP space is not well-suited for the current text-to-motion dataset. The CLIP text encoder is not effective at distinguishing between text descriptions that describe different motions, and the MotionCLIP's motion encoder, built upon the CLIP space, struggles even more to differentiate between different motions.

Therefore, in the end, we selected TMR as our alignment space. Compared to MotionCLIP, the features extracted by its text encoder and motion encoder have satisfactory differentiation in the latent space. Experiments have also demonstrated that it is highly suitable for our downstream tasks.

\section{The choice of diffusion model}

Using diffusion directly instead of the latent diffusion can lead to more applications, such as joint trajectory control~\cite{shafir2023human,wan2023tlcontrol}. However, we ultimately chose latent diffusion as our paradigm, not only for performance enhancement but also based on the reason outline below.

\begin{figure}[t]
   \centering
   \subfloat[Before RVQVAE]{
   \begin{minipage}{0.5\linewidth}
       \centering
       \includegraphics[width=\linewidth]{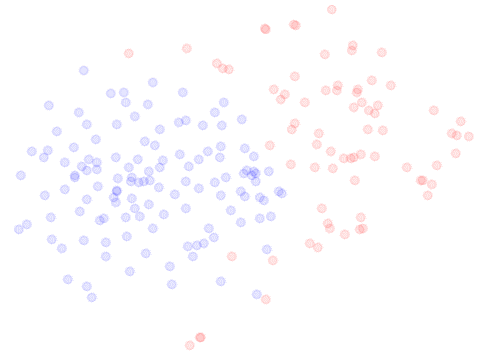}
   \end{minipage}}
   \subfloat[After RVQVAE]{
   \begin{minipage}{0.5\linewidth}
       \centering
       \includegraphics[width=\linewidth]{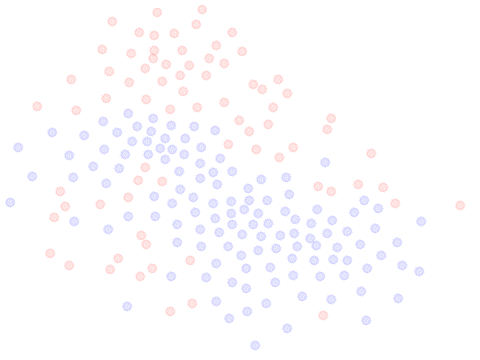}
   \end{minipage}}
   \caption{The t-SNE visualizations of motions before and after RVQVAE (blue: motions in BEATX, red: motions in AMASS). 
   }
   \label{fig:tsne}
\end{figure}

Due to the distinct differences between the BEATX and AMASS datasets, with BEATX focusing on fine grained hand movements and full-body motions synchronized with speech rhythm, and AMASS emphasizing a wide variety of rich full-body actions, Figure~\ref{fig:tsne} is a direct visual comparison using t-SNE dimensionality reduction reveals significant disparities between the two datasets. 

Based on this observation, instead of directly mixing the two datasets for training, we first train an autoencoder to map the two distinct datasets into the same latent space, thereby reducing their data variability.

\section{Additional Experimental Results}

\subsection{Training strategy}

In the main paper, we use implicit labels to provide corresponding text features for all motions. This approach allows each motion to have an associated text feature even in the absence of a corresponding text label. Specifically, we achieve this by using the motion encoder, which was constructed in the previous step to build the text-motion alignment space, to extract motion features. These motion features are considered equivalent to the corresponding text features..

An obvious observation is that, although text labels are missing in the speech-to-motion dataset, the text-to-motion dataset does contain paired text labels. Since in inference stage, we generally use text to express the user's intent, and also the evaluation metrics are also based on text, would it be feasible during training to directly use the text labels through the text encoder for subsequent training instead of motion? Could this lead to a performance improvement?

As shown in Table~\ref{tab:text}, we conducted this experiment, but the conclusion was that using motion through the motion encoder directly yielded better results than using text through the text encoder. In Table~\ref{tab:text}, "using 1/3 text" means that for motions with text descriptions, only one-third of the data uses text through text encoder to extract features through, while the rest use motion through motion encoder to extract features. The more text we use, the worse results we get. One possible reason is that our training process consistently used fixed 128 frames, approximately 4 seconds long. However, the HumanML3d text descriptions vary in length, ranging from 2 to 10 seconds. For instance, for a text label like "person walking with their arms swinging back to front and walking in a general circle. walking in a general circle.", it describes a 10-second motion, whereas our training process randomly sampled 4-second segments. If the first 4 seconds are sampled, the subsequent "walking in a general circle." action segment would be truncated. This difference in description between text and motion could lead to performance degradation compared to directly using motion throughout the training process.

\subsection{More Results for Main Paper Figure 4 and 5}
\begin{figure*}[!t]
  \centering
  \includegraphics[width=\linewidth]{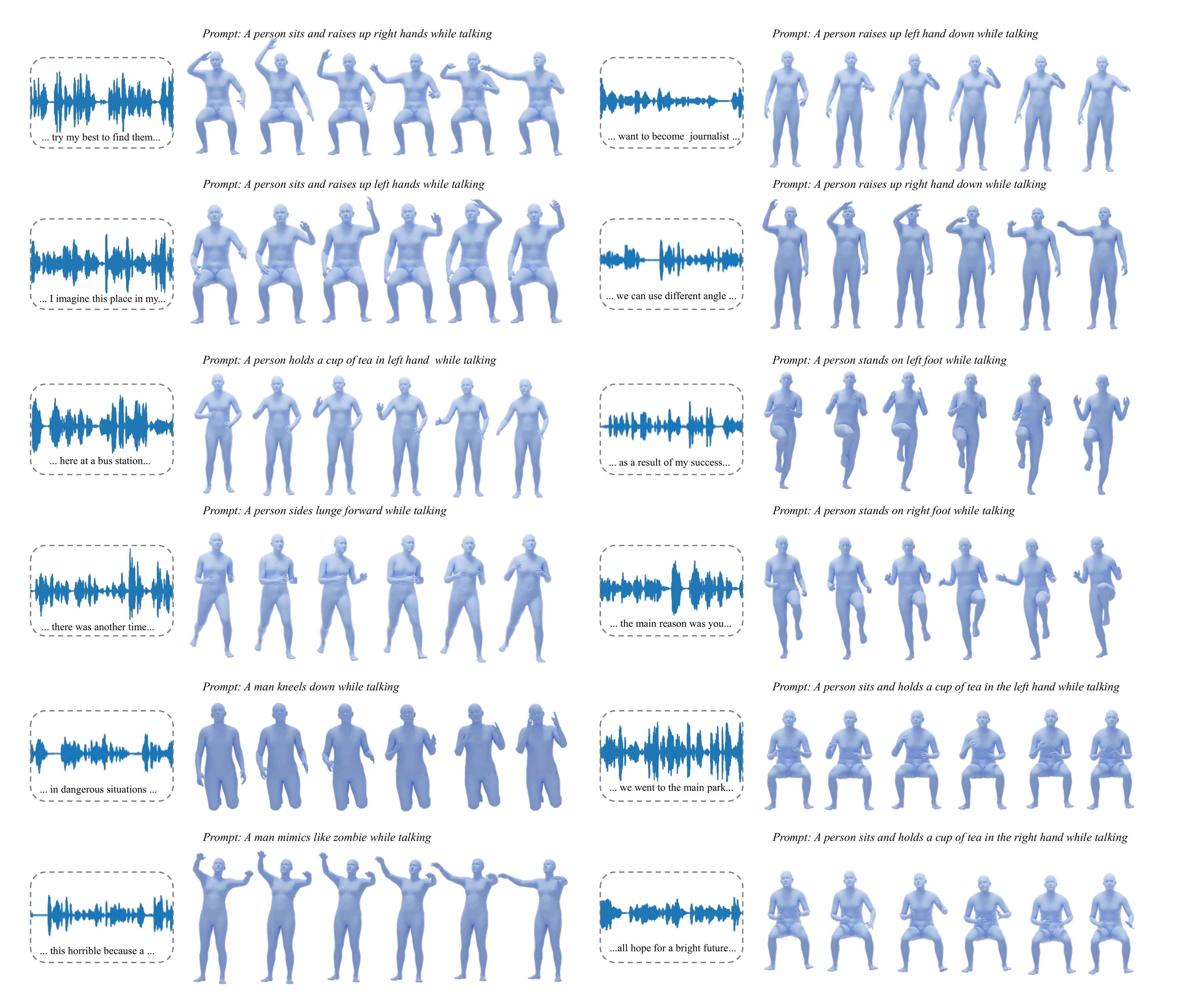}
  \caption{More results for main paper Figure 4 (full-body
synergistic generation).
  }
  \label{qualitative}
\end{figure*}

\begin{figure*}[!t]
  \centering
  \includegraphics[width=\linewidth]{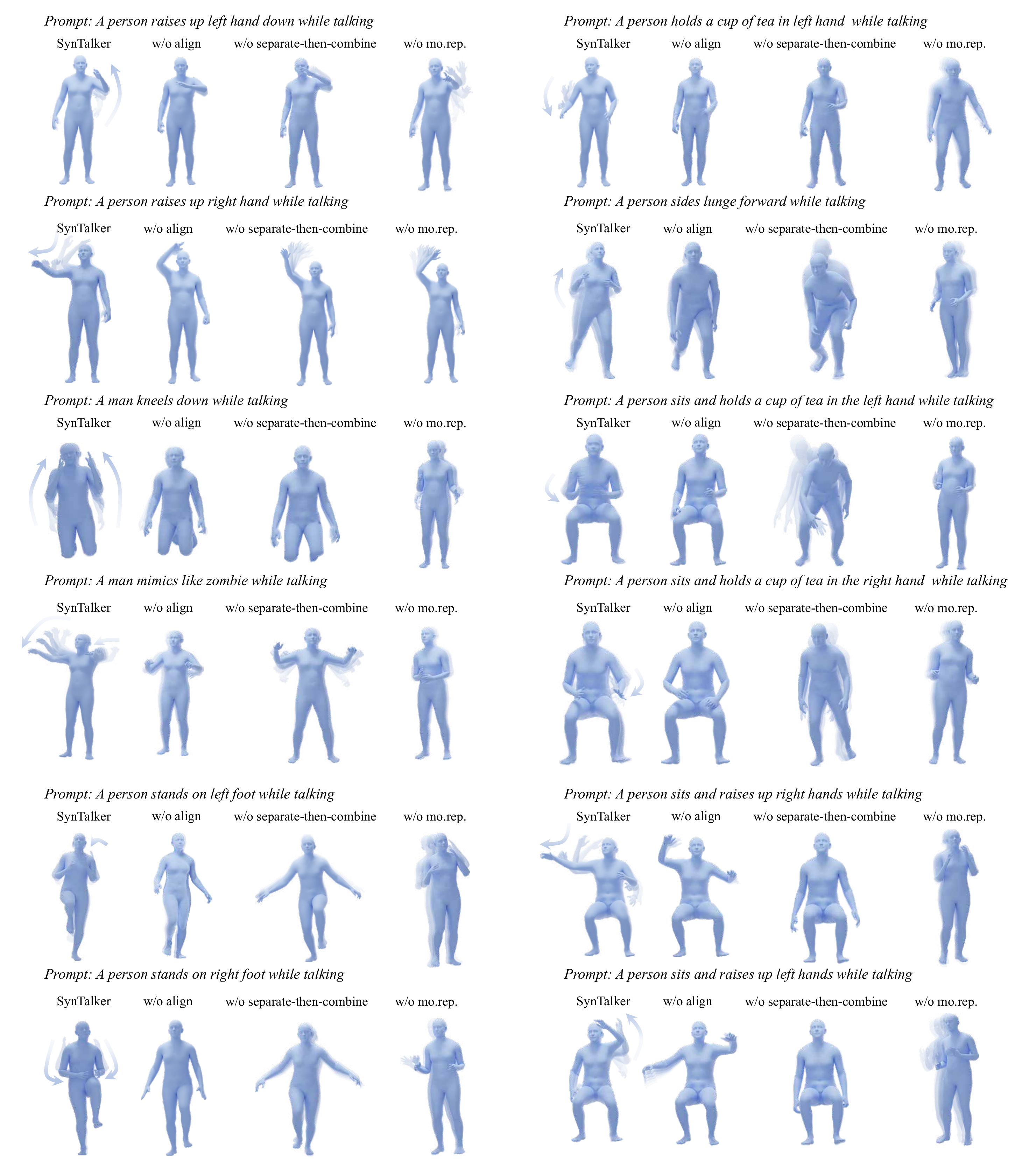}
  \caption{More results for main paper Figure 5 (ablation study)
  }
  \label{ablation}
\end{figure*}

To compensate main paper Figure 4, we provide more results of full-body synergistic generation in Figure~\ref{qualitative}, which illustrates the idealized generation quality of our methods under various of audio and prompt conditions. Furthermore, to compensate main paper Figure 5, we provide more results of ablation study in Figure~\ref{ablation}, showing that the ablation results are solid so that our training and inference components are solid.

\subsection{Comparison with FreeTalker and GestureDiffuCLIP}

\textbf{FreeTalker}

FreeTalker~\cite{yang2024Freetalker} is a neural model designed to perform audio-to-motion and text-to-motion tasks simultaneously within the same network. We experimented with feeding both speech and text inputs to the model simultaneously during inference. As shown in Figure~\ref{fig:compare_ft}, the model fails to generate synergistic co-speech motion when receiving both conditions simultaneously. This highlights that training the model directly for audio-to-motion and text-to-motion tasks does not automatically enable it to produce synergistic results under dual conditions.

\noindent \textbf{GestureDiffuCLIP} 

\begin{figure*}[t]
    \centering
    \includegraphics[width=.7\linewidth]{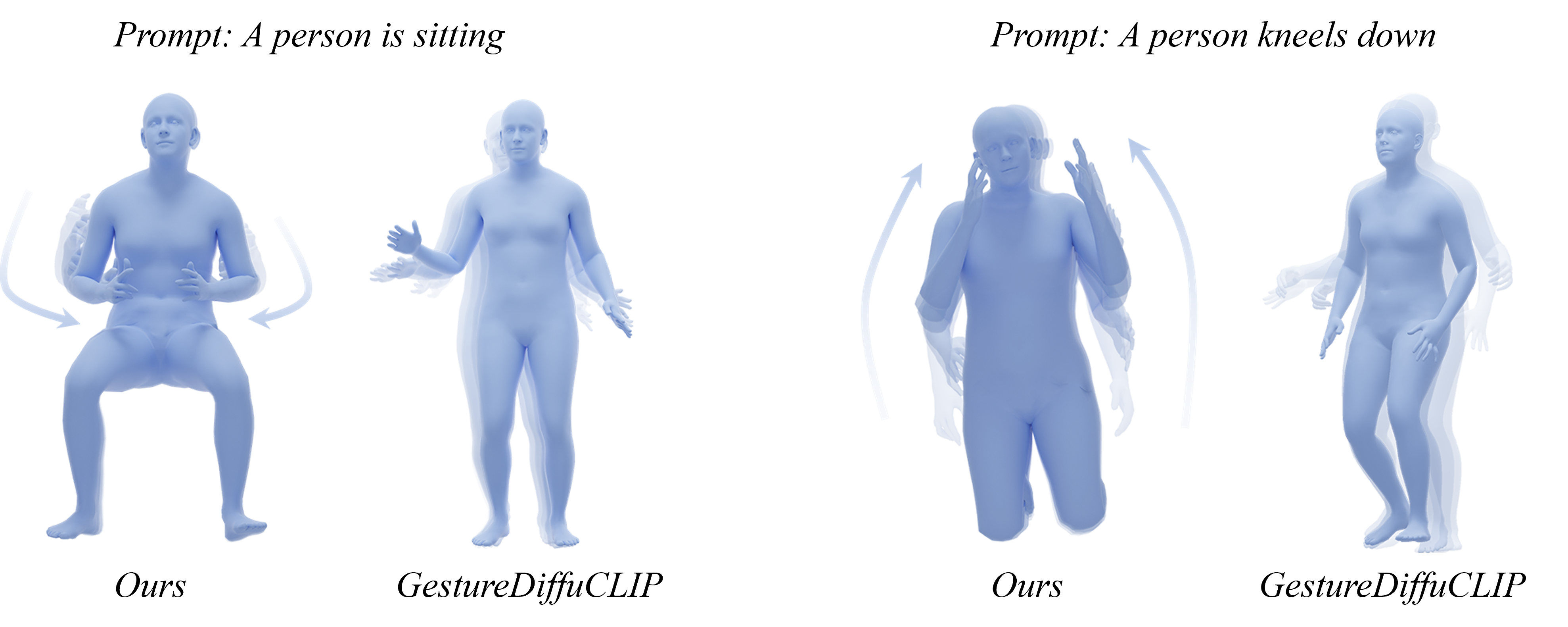}
    \caption{Qualitative comparison with GestureDiffuCLIP conditioned on audio and text features simultaneously. }
    \label{fig:compare_gdc}
\end{figure*}

\begin{figure*}[t]
    \centering
    \includegraphics[width=.7\linewidth]{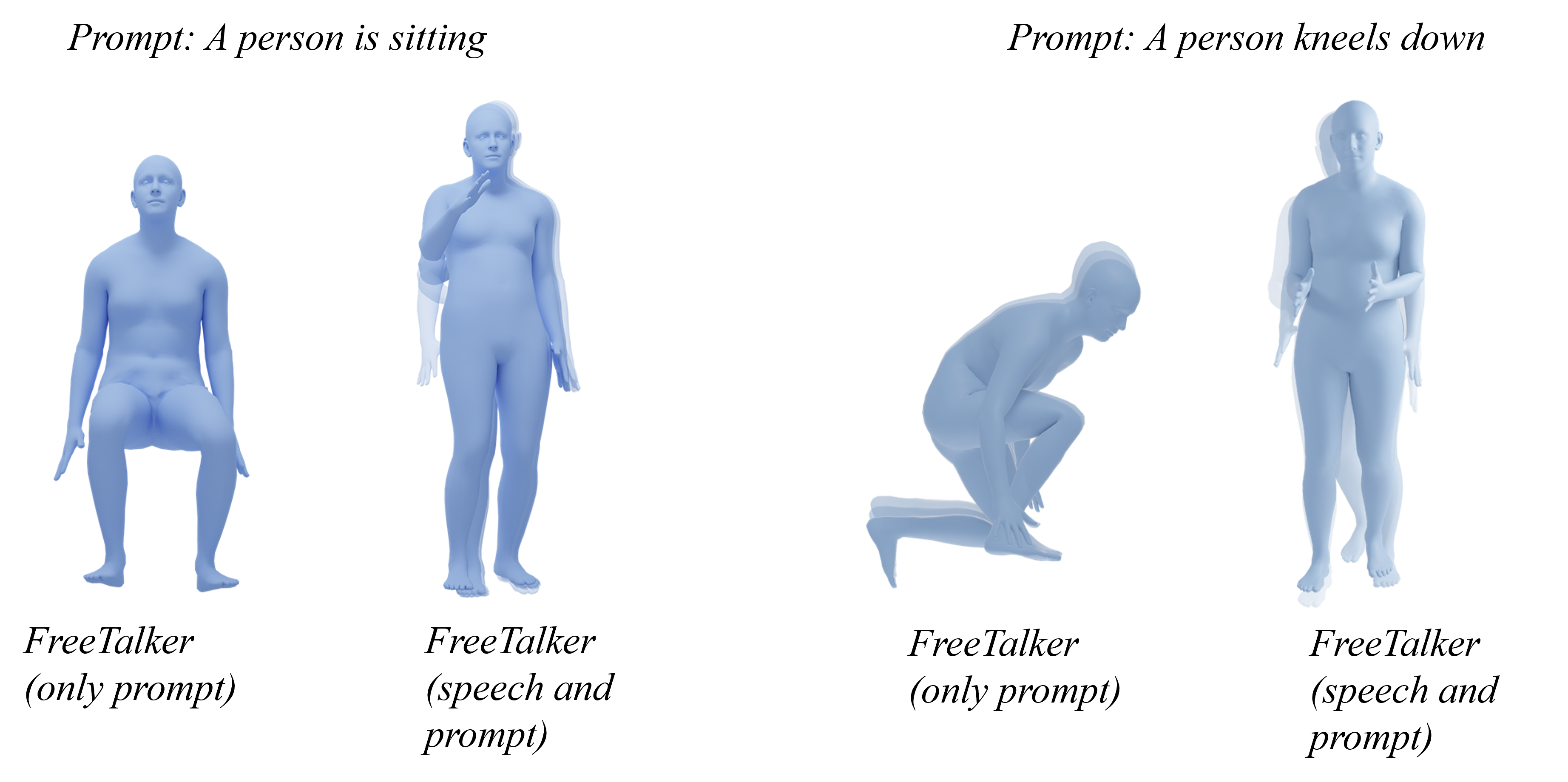}
    \caption{Qualitative results of FreeTalker conditioned on audio and text features simultaneously}
    \label{fig:compare_ft}
\end{figure*}

GestureDiffuCLIP~\cite{Ao2023GestureDiffuCLIP} enables users to control the style of generated co-speech motion using text prompts. To illustrate the fundamental difference between our synergistic co-speech motion generation and text-based style control, we conducted a simple comparison shown in Figures~\ref{fig:compare_gdc}. As the authors did not release the source code, we re-implemented the model for this comparison. As depicted in the figures, under the textual prompts "A person is sitting while talking" and "A person kneels down while talking," our results (left) accurately follow the prompts, while GestureDiffuCLIP's (right) only shows a tendency to squat, underscoring the fundamental difference between text-based style control and text-based motion control.

\subsection{Remark on User Study}

We had considered adding a user study to further enhance the completeness of our work, but we found it difficult to carry out for the following reasons:

To the best of our knowledge, the only work similar to ours is GestureDiffuCLIP~\cite{Ao2023GestureDiffuCLIP}. However, this work has not been open-sourced, and using our own reproduced version for quantitative comparison would be inherently unfair. Additionally, due to the limitations of its training dataset, it is almost entirely incapable of generating any lower body movements.

Other open-source works, such as FreeTalker~\cite{yang2024Freetalker}, have already been compared in the experiments mentioned above. These methods do not support simultaneously processing speech and text prompts as inputs.

\begin{table*}[t]
    \centering
    \caption{\textbf{Comparison with the state-of-the-art methods on HumanML3D~\cite{Guo_2022_CVPR} test set.} We compute standard metrics following~\cite{Guo_2022_CVPR}. For each metric, we repeat the evaluation 20 times and report the average with 95\% confidence interval.}
    
    {
    \begin{tabular}{l c c c c c c c}
    \toprule
    \multirow{2}{*}{Methods}  & \multicolumn{3}{c}{R-Precision $\uparrow$} & \multirow{2}{*}{FID $\downarrow$} & \multirow{2}{*}{MM-Dist $\downarrow$} & \multirow{2}{*}{Diversity $\uparrow$}\\

    \cline{2-4}
    ~ & Top-1 & Top-2 & Top-3 \\
    
    \midrule

        \ourapproach & \et{0.375}{.003} & \et{0.564}{.003} & \et{0.681}{.002} & \et{4.385}{.034} & \et{4.499}{.012} & \et{9.374}{.073}  \\

        \ourapproach\;(use MotionCLIP)& \et{0.144}{.002} & \et{0.253}{.002} & \et{0.345}{.003} & \et{13.403}{.087} & \et{6.552}{.013} & \et{7.984}{.083} \\

        \ourapproach\;(w/o text-to-motion alignment) & \et{0.429}{.003} & \et{0.622}{.004} & \et{0.732}{.004} & \et{0.509}{.013} & \et{4.033}{.013} & \et{10.231}{.096}\\ 

        \ourapproach\;(use 1/1 text) & \et{0.321}{.002} & \et{0.502}{.002} & \et{0.617}{.002} & \et{7.046}{.044} & \et{5.007}{.011} & \et{8.733}{.075}  \\

        \ourapproach\;(use 1/2 text) & \et{0.336}{.002} & \et{0.517}{.003} & \et{0.631}{.003} & \et{6.388}{.046} & \et{4.909}{.012} & \et{8.872}{.075}  \\
 
        \ourapproach\;(use 1/3 text) & \et{0.349}{.002} & \et{0.530}{.002} & \et{0.641}{.003} & \et{5.983}{.051} & \et{4.764}{.012} & \et{8.963}{.076}  \\

        \ourapproach\;(use 1/4 text) & \et{0.360}{.002} & \et{0.549}{.003} & \et{0.658}{.004} & \et{5.227}{.043} & \et{4.658}{.012} & \et{9.118}{.075}  \\

    \bottomrule
    \end{tabular}
    }
    \vspace{1mm}
    \label{tab:text}

\end{table*}

\end{document}